%% file: main.tex
\documentclass[11pt,a4paper]{article}
\usepackage[letterpaper]{geometry}
\usepackage[hyperref]{naaclhlt2019}
\usepackage{naaclhlt2019}
\usepackage{times}
\usepackage{latexsym}
\usepackage{amsmath}
\usepackage{multirow}
\usepackage{url}
\usepackage{xcolor}
\usepackage{bbm}
\usepackage{graphicx}
\usepackage{tabularx, booktabs}
\makeatletter
\newcommand{\@emptybiblabel}[1]{}
\makeatother
\usepackage{marginnote}

\usepackage{bbding}
\usepackage{pifont}
\usepackage{wasysym}
\usepackage{amssymb}
\newcommand{\Yes}{\checkmark}
\newcommand{\No}{$\times$}
\newcommand{\wsq}{PAWS$_\textrm{QQP}$}
\newcommand{\wsw}{PAWS$_\textrm{Wiki}$}
\newcommand{\wswswap}{PAWS$_\textrm{Wiki-Swap}$}
\newcommand{\wsqw}{PAWS$_\textrm{QQP+Wiki}$}
\newcommand{\qqp}{\textsc{QQP}}
\newcommand{\paws}{PAWS}

\newcommand{\specialcell}[2][c]{%
  \begin{tabular}[#1]{@{}c@{}}#2\end{tabular}}

\usepackage{linguex}

\aclfinalcopy

\title{PAWS: Paraphrase Adversaries from Word Scrambling}

\author{Yuan Zhang ~~~~ Jason Baldridge ~~~~ Luheng He \\
  Google AI Language \\
  {\tt \{zhangyua,jridge,luheng\}@google.com}}

\date{}

\begin{document}
\maketitle
\begin{abstract}
\input{abstract}

\end{abstract}

\input{intro}

\input{related}

\input{example_gen}

\input{ws_dataset}

\input{models}

\input{experiments}

\input{conclusion}

\section*{Acknowledgement}
We would like to thank our anonymous reviewers and the Google AI Language team, especially Emily Pitler, for the insightful comments that contributed to this paper. Many thanks also to the Data Compute team, especially Ashwin Kakarla and Henry Jicha, for their help with the
annotations

\bibliography{naaclhlt2019}
\bibliographystyle{acl_natbib}

\end{document}

%% file: abstract.tex
Existing paraphrase identification datasets lack sentence pairs that have high lexical overlap without being paraphrases. Models trained on such data fail to distinguish pairs like \textit{flights from New York to Florida} and \textit{flights from Florida to New York}. This paper introduces PAWS (\textbf{P}araphrase \textbf{A}dversaries from \textbf{W}ord \textbf{S}crambling), a new dataset with 108,463 well-formed paraphrase and non-paraphrase pairs with high lexical overlap. Challenging pairs are generated by controlled word swapping and back translation, followed by fluency and paraphrase judgments by human raters. State-of-the-art models trained on existing datasets have dismal performance on PAWS ($<$40\% accuracy); however, including PAWS training data for these models improves their accuracy to 85\% while maintaining performance on existing tasks. In contrast, models that do not capture non-local contextual information fail even with PAWS training examples. As such, PAWS provides an effective instrument for driving further progress on models that better exploit structure, context, and pairwise comparisons.

%% file: intro.tex
\input{tables/intro_examples_table}

\section{Introduction} \label{sec:intro}

Word order and syntactic structure have a large impact on sentence meaning. Even small perturbation in word order can completely change interpretation. Consider the following related sentences. %

\ex. \label{ex:1} Flights from New York to Florida.

\vspace{-0.2in}
\ex. \label{ex:2} Flights to Florida from NYC.

\vspace{-0.2in}
\ex. \label{ex:3} Flights from Florida to New York.

\noindent
All three have high bag-of-words (BOW) overlap. However, \ref{ex:2} is a paraphrase of \ref{ex:1}, while \ref{ex:3} has a very different meaning from \ref{ex:1}.

\begin{figure}[t]
\centering
\includegraphics[trim={2.8in 1.8in 0.0in 1in},clip,width=0.5\textwidth]{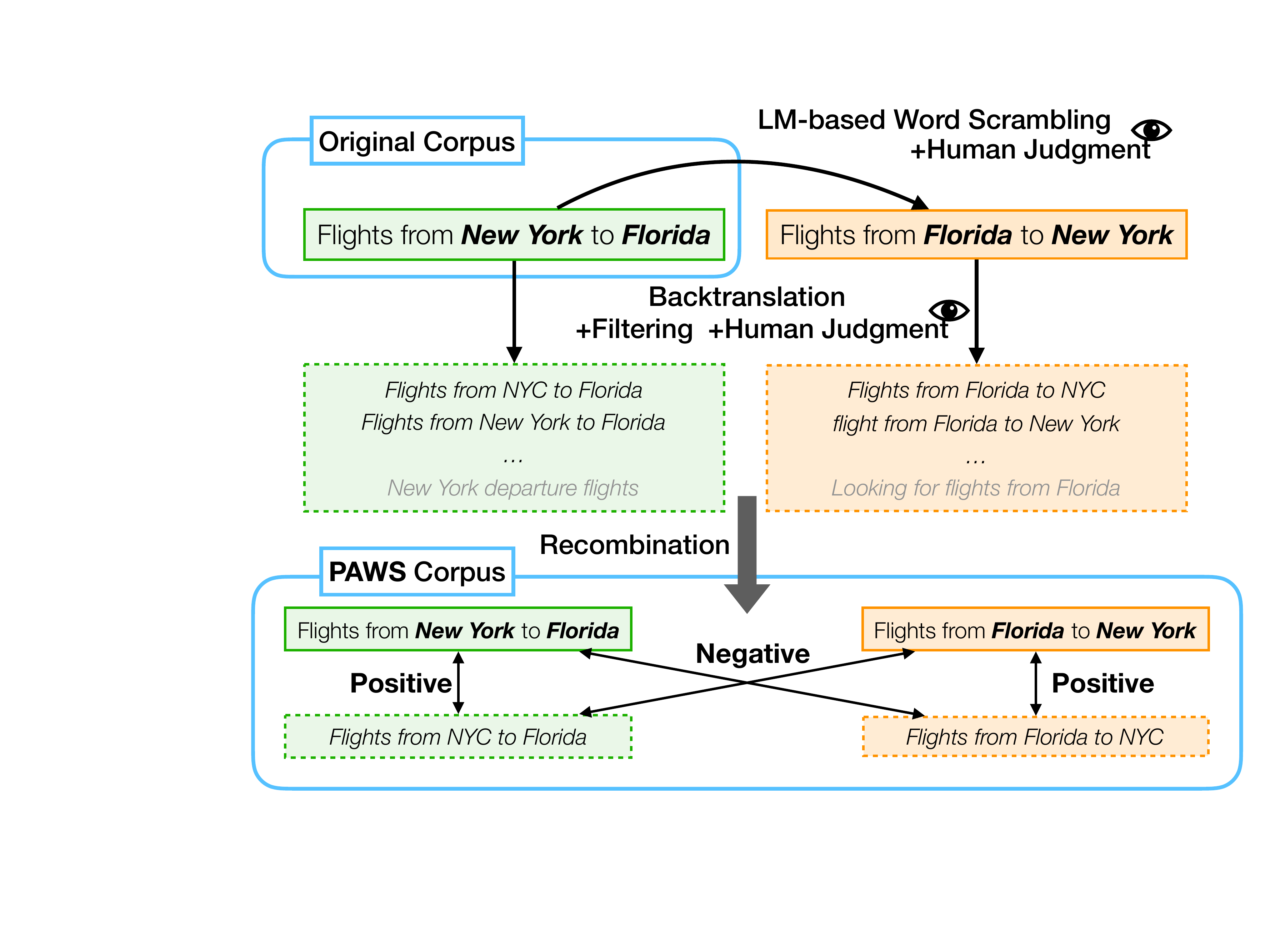}
\caption{\paws\ corpus creation workflow.}\label{fig:wsflow}
\end{figure}

Existing datasets lack non-paraphrase pairs like \ref{ex:1} and \ref{ex:3}. The Quora Question Pairs (\qqp) corpus contains 400k real world pairs, but its negative examples are drawn primarily from related questions. Few have high word overlap, and of the $\sim$1,000 pairs with the same BOW, only 20\% are not paraphrases. This provides insufficient representative examples to evaluate models' performance on this problem, and there are too few examples for models to learn the importance of word order. Table \ref{tab:motivating_examples} shows that models trained on \qqp\ are inclined to mark any sentence pairs with high word overlap as paraphrases despite clear clashes in meaning. Models trained or evaluated with only this data may not perform well on real world tasks where such sensitivity is important.

To address this, we introduce a workflow (outlined in Figure \ref{fig:wsflow}) for generating pairs of sentences that have high word overlap, but which are balanced with respect to whether they are paraphrases or not. Using this process, we create  \paws\ (\textbf{P}araphrase \textbf{A}dversaries from \textbf{W}ord \textbf{S}crambling), a dataset constructed from sentences in Quora and Wikipedia. Examples are generated from controlled language models and back translation, and given five human ratings each in both phases. A final rule recombines annotated examples and balances the labels. Our final \paws\ dataset will be released publicly with 108,463 pairs at \url{https://g.co/dataset/paws}.

We show that existing state-of-the-art models fail miserably on \paws\ when trained on existing resources, but some perform well when given \paws\ training examples. BERT \cite{devlin2018bert} fine-tuned on \qqp\ achieves over 90\% accuracy on \qqp, but only 33\% accuracy on \paws\ data in the same domain. However, the accuracy on \paws\ boosts to 85\% by including 12k \paws\ training pairs (without reducing \qqp\ performance). Table \ref{tab:motivating_examples} also shows that the new model is able to correctly classify challenging pairs. Annotation scale is also important: our learning curves show strong models like BERT improve with tens of thousands of training examples. %

Our experimental results also demonstrate that \paws\ effectively measures sensitivity of models to word order and structure. Unlike BERT, a simple BOW model fails to learn from \paws\ training examples, demonstrating its weakness at capturing non-local contextual information. Our experiments show that the gains from \paws\ examples correlate with the complexity of models.

%% file: tables/intro_examples_table.tex
\begin{table*}[t!]
\centering
\footnotesize
\setlength{\tabcolsep}{4pt}
\begin{tabularx}{\linewidth}{c@{~~} p{2.05in} p{2.05in} cccc}
\toprule
& \multirow{2}{*}{Sentence 1} & \multirow{2}{*}{Sentence 2} & \multirow{2}{*}{Gold} & \multirow{2}{*}{BOW} & \multirow{2}{*}{BERT} & BERT+ \\
& & & & & & \paws \\
\midrule
(1) & Can a bad person become good?	&
    Can a good person become bad? &
    \textbf{N} & Y & Y & \textbf{N} \\
(2) & Which is the cheapest flight from anywhere in South America to Europe?
    & Which is the cheapest flight from anywhere in Europe to South America?
    & \textbf{N} & Y & \textbf{N} & \textbf{N} \\
(3) & ``Taunton Castle'' was on August 1 in Rio de Janeiro and on October 31 in Penang.
    & ``Taunton Castle'' was at Penang on 1 August and Rio de Janeiro on 31 October. &
    \textbf{N} & Y & Y & \textbf{N} \\
(4) & Although interchangeable, the body pieces on the 2 cars are not similar. &
    Although similar, the body parts are not interchangeable on the 2 cars. &
    \textbf{N} & Y & Y & \textbf{N} \\
(5) & Katz was born in Sweden in 1947 and moved to New York City at the age of 1. &
    Katz was born in 1947 in Sweden and moved to New York at the age of one. &
    \textbf{Y} & \textbf{Y} & \textbf{Y} & \textbf{Y} \\
(6) & It was not the sales manager who hit the bottle that day, but the office worker with the serious drinking problem. &
    That day the office manager, who was drinking, hit the problem sales worker with a bottle, but it was not serious. &
    \textbf{N} & Y & Y & \textbf{N}\\
\bottomrule
\end{tabularx}
\caption{Paraphrase/Non-paraphrase (Y/N) pairs with high bag-of-words (BOW) overlap. (1)-(5) are drawn from \paws, and (6) is from \newcite{MitchellL08}. Both a simple BOW and the state-of-the-art BERT \cite{devlin2018bert} models, if trained/fine-tuned on the Quora Question Pairs (\qqp) corpus, classify all of them (with one exception) as paraphrases (Y). A BERT model fine-tuned on both \qqp\ and \paws\ examples (BERT+\paws), however, is able to get them correct.}\label{tab:motivating_examples}
\end{table*}

%% file: related.tex
\section{Related Work} \label{sec:related}

Existing data creation techniques have focused on collecting paraphrases, e.g. from co-captions for images \cite{Lin:14}, tweets with shared URLs \cite{LanQHX17}, subtitles \cite{Creutz18}, and back translation \cite{IyyerWGZ18}. Unlike all previous work, we emphasize the collection of challenging negative examples.

Our work closely relates to the idea of crafting adversarial examples to break NLP systems. Existing approaches mostly focused on adding label-preserving perturbations to inputs, but with the effect of distracting systems from correct answers. Example perturbation rules include adding noise to inputs \cite{jia2017adversarial,chen2018attacking}, word replacements \cite{alzantot2018generating,ribeiro2018semantically}, and syntactic transformation \cite{IyyerWGZ18}. A notable exception is \citet{glockner2018breaking}: they generated both entailment and contradiction examples by replacing words with their synonyms or antonyms. Our work presents two main departures. We propose a novel method that generates challenging examples with balanced class labels and more word reordering variations than previous work. In addition, we release to public a large set of 108k example pairs with high-quality human labels. We believe the new dataset will benefit future research on both adversarial example generation and improvement of model robustness.

In our work, we demonstrate the importance of capturing non-local contextual information in the problem of paraphrase identification. This relates to prior work on probing sentence representations for their linguistic properties, such as how much syntactic information is encoded in representations \cite{conneau2018what,tenney2019what,ettinger2018assessing}. There also exists prior work that directly uses structural information in modeling \cite{filice2015structural,liu2018structured}. All these prior approaches were evaluated on existing datasets. In contrast, we perform studies on PAWS, a new dataset that emphasizes the importance of capturing structural information in representation learning. While developing new models is beyond the scope of this paper, this new dataset can facilitate research in this direction.

%% file: example_gen.tex
\begin{table*}[t!]
\centering
\footnotesize
\setlength{\tabcolsep}{4pt}
\begin{tabularx}{\linewidth}{c@{~~} p{2.25in} p{2.25in} l}
\toprule
& Sentence 1 & Sentence 2 & Generation Type \\
\midrule
(1) & Can a \textbf{bad} person become \textbf{good}?	&
    Can a \textbf{good} person become \textbf{bad}? &
    Adjective swap \\
(2) & \textbf{Jerry} looks over \textbf{Tom}'s shoulder and gets punched. &
    \textbf{Tom} looks over \textbf{Jerry}'s shoulder and gets punched. &
    Named entity swap \\
(3) & The team also toured in Australia \textbf{in 1953}. &
    \textbf{In 1953,} the team also toured in Australia. &
    Temporal phrase swap \\
(4) & Erikson \textbf{formed} the rock band Spooner with two fellow musicians. &
    Erikson \textbf{founded} the rock band Spooner with two fellow musicians. &
    Word replacement \\
    
\bottomrule
\end{tabularx}
\caption{Examples of typical types of generation. (1) and (2) are from the word swapping method, while (3) and (4) are from the back translation method. Boldface indicates changes in each example. }\label{tab:generation_type}
\end{table*}

\section{\paws\ Example Generation} \label{sec:exgen}

We define a \textbf{\paws\ pair} to be a pair of sentences with high bag-of-words (BOW) overlap but different word order. In the Quora Question Pairs corpus, 80\% of such pairs are paraphrases. Here, we describe a method to automatically generate non-trivial and well-formed \paws\ pairs from real-world text in any domain (this section), and then have them annotated by human raters (Section \ref{sec:ws_dataset}).

Our automatic generation method is based on two ideas. The first swaps words to generate a sentence pair with the same BOW, controlled by a language model. The second uses back translation to generate paraphrases with high BOW overlap but different word order. These two strategies generate high-quality, diverse \paws\ pairs, balanced evenly between paraphrases and non-paraphrases.

\begin{figure}[t]
\centering
\includegraphics[width=0.48\textwidth]{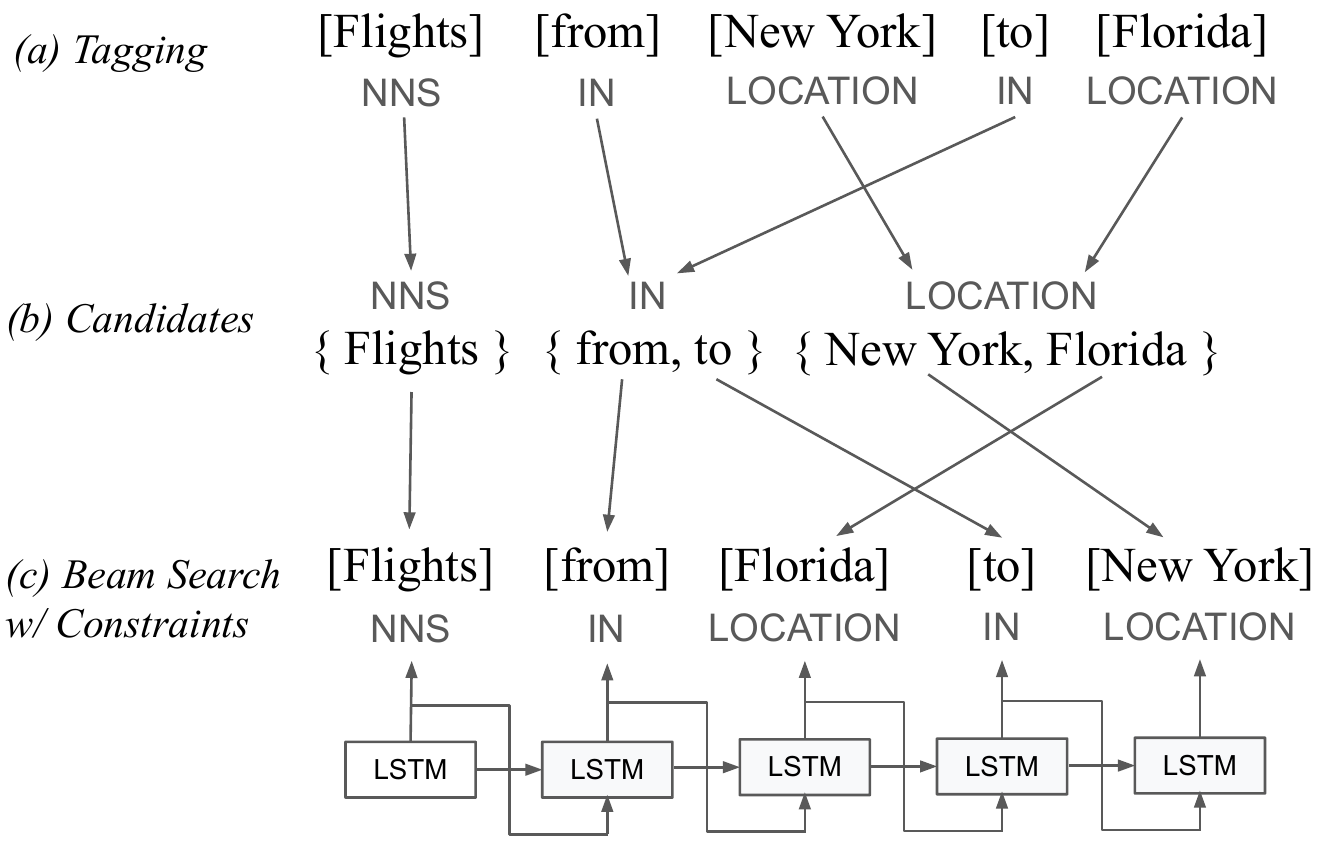}
\caption{Illustration of the generation method in three steps. (a) Tag words and phrases with part-of-speech (POS) and named entities. (b) Build candidate sets by grouping words and phrases with the same tag. (c) Under the constraints of tag sequence template and candidate sets, find sentences with high language model scores using beam search. 
}\label{fig:method}
\end{figure}

\subsection{Word Swapping}

Our first phase generates well-formed sentences by swapping words in real world text. Most text generation models rely on large amount of training data \cite{IyyerWGZ18,GuuHOL18,GuptaASR18,LiJSL18}, which is unfortunately not available in our case. We thus propose a novel generation method based on language modeling and constrained beam search. The goal is to find a sentence that achieves high language model score as well as satisfying all constraints. High scores indicate that generated sentences are natural and well-formed, and constraints ensure generated pairs have the same BOW. 

Figure \ref{fig:method} illustrates the generation procedure. First, given an input sentence, a CRF-based part-of-speech tagger tags each word. We further detect person names, locations, and organizations using a named entity recognizer, and replace POS with entity tags if probability scores are above 95\%.\footnote{We pick this threshold to achieve about 95\% precision.} The sequence of tags of words and phrases form a template for the input. 

Our beam search method then fills in each slot of the template from left to right, scoring each state by a language model trained on one billion words \cite{ChelbaMSGBKR14}. The candidate words and phrases for each slot are drawn from the input based on its tag. In Figure \ref{fig:method}, for example, the second slot must be filled with a \textsc{Location} from two candidate \textit{New York} and \textit{Florida}. Candidates are drawn without replacement so the generated sentence and the input have exactly the same bag-of-words. Note that this template-based constraint is more restrictive than the BOW requirement, but we choose it because it significantly reduces the search space. With this constraint, the method achieves high generation quality without a large beam. In practice, beam size is set to 100, which produces near-optimal results in most cases.

Let $s'$ be the best sentence in the beam other than the input sentence $s$, and $LM(\cdot)$ be their log-likelihood by the language model. We take $(s, s')$ as a good word-swapping pair if $LM(s') \ge LM(s) - t$.\footnote{In a preliminary stage, we noticed that many pairs were simply a permutation of a list, like ``A and B'' changed to ``B and A''. For the diversity of the dataset, 99\% of these are pruned via hand-crafted, heuristic rules.} We manually pick the threshold $t{=}3.0$ for a good balance between generation quality and coverage. Examples (1) and (2) in Table \ref{tab:generation_type} are representative examples from this generation method.

\subsection{Back Translation} \label{sec:backtrans}

Because word order impacts meaning, especially in English, the swapping method tends to produce non-paraphrases. Our preliminary results showed that the distribution of paraphrase to non-paraphrases from this method is highly imbalanced (about 1:4 ratio). However, we seek to create a balanced dataset, so we use an additional strategy based on back translation---which has the opposite label distribution and also produces greater diversity of paraphrases while still maintaining a high BOW overlap.

The back translation method takes a sentence pair and label $(s_1, s_2, l)$ as input. For each sentence, the top-$k$ translations are obtained from an English-German neural machine translation model (NMT); then each of these is translated back to English using another German-English NMT model, providing a resulting top-$k$ results. We chose German as the pivot language because it produced more word reordering variations than other languages and the translation quality was good. Both models have the same architecture \cite{wu2016google} and are trained on WMT14. This results in $k^2$ back translations before deduplication. We chose $k{=}5$. To obtain more pairs with the \paws\ property, we further filter back translations by their BOW similarities to the input and their word-order inversion rates, as described below.

We define BOW similarity as the cosine similarity $\alpha$ between the word count vectors of a sentence pair. Pairs generated from the swapping strategy have score $\alpha=1.0$, but here we relax the threshold to 0.9 because it brings more data diversity and higher coverage, while still generating paraphrases of the input with high quality.

\begin{figure}[t]
\centering
\includegraphics[width=0.45\textwidth]{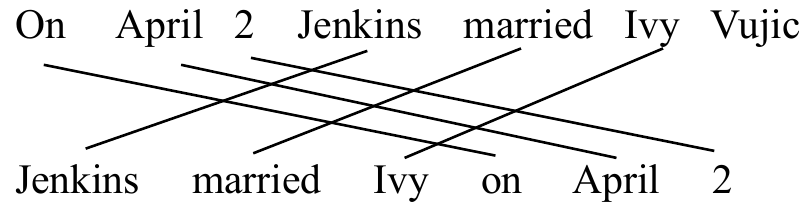}
\caption{An example of how to compute inversion rate.}\label{fig:inverse_ratio}
\end{figure}

To define the word-order inversion rate, we first compute word alignments between a sentence pair in a heuristic way by assuming they are one-to-one mapping and are always monotonic. For example, if the first sentence has three instances of \textit{dog} and the second has two, we align the first two instances of \textit{dog} in the same order and skip the third one. The inversion rate is then computed as the ratio of cross alignments. Figure \ref{fig:inverse_ratio} is an example pair with six alignments. There are 15 alignment pairs in total and 9 of them are crossed, e.g. alignments of \textit{on} and \textit{married}. The inversion rate of this example is therefore $9/15=0.6$.
We sample back translation results such that at least half of the pairs have inversion rate over 0.02; this way, the final selected pairs cover interesting transformations of both word-order changes and word replacement. Examples (3) and (4) in Table \ref{tab:generation_type} are representative examples from back translation.

\paragraph{Label Balancing} Figure \ref{fig:wsflow} illustrates the process of constructing the final label-balanced set based on human annotations. The set first includes all pairs from back translation, which are mostly paraphrases. For each labeled pair $(s_1, s_2)$ from swapping and a labeled pair $(s_1, s_1')$ from back translation, the set further includes the pair $(s_2, s_1')$ based on the rules: (1) $(s_2, s_1')$ is paraphrase if both $(s_1, s_2)$ and $(s_1, s_1')$ are paraphrases; (2) $(s_2, s_1')$ is non-paraphrase if exactly one of $(s_1, s_2)$ and $(s_1, s_1')$ is non-paraphrase; (3) otherwise $(s_2, s_1')$ is not included because its label is unknown. We also consider pairs $(s_2', s_1)$ and $(s_2', s_1')$ in the similar way if $s_2'$ is a back translation of $s_2$ with human labels.

%% file: ws_dataset.tex
\section{PAWS Dataset} \label{sec:ws_dataset}

Using the example generation strategies described in Section \ref{sec:exgen} combined with human paraphrase annotations, we create a large new dataset, \textbf{PAWS} that contains both paraphrase and non-paraphrase pairs that have both high bag-of-words overlap and word reordering. Source sentences are drawn from both the Quora Question Pairs (\qqp) corpus \cite{iyer2017first} and Wikipedia.\footnote{\url{https://dumps.wikimedia.org}} From these, we produce two datasets, \wsq\ and \wsw.

\begin{table}[t]
    \centering
    \begin{tabular}{l@{~~}rr}
         & Quora & Wikipedia \\
    \midrule
        \# Raw pairs & 16,280 & 50,000 \\
    \midrule
        \textit{Sentence correction} & & \\
        ~~~~\# Accepted pairs & 10,699 & 39,903 \\
        ~~~~\# Fixed pairs & 3,626 & 7,387 \\
        ~~~~\# Rejected pairs & 1,955 & 2,710 \\
    \midrule
        \textit{Paraphrase identification} & & \\
        ~~~~Total \# pairs & 14,325 & 47,290 \\
        ~~~~~~~~paraphrase & 4,693 & 5,725 \\
        ~~~~~~~~non-paraphrase & 9,632 & 41,565 \\
        ~~~~Human agreement & 92.0\% & 94.7\% \\
    \midrule
        \textit{After post-filtering} & & \\
        ~~~~Total \# pairs & 12,665 & 43,647 \\
        ~~~~Human agreement & 95.8\% & 97.5\% \\
    \end{tabular}
    \caption{Detailed counts for examples created via the swapping strategy, followed by human filtering and paraphrase judgments.}
    \label{tb:swap_annotation}
\end{table}

We start by producing swapped examples from both \qqp\ and Wikipedia. Both sources contain naturally occurring sentences covering many topics. On both corpora only about 3\% of candidates are selected for further processing---the rest are filtered because there is no valid generation candidate that satisfies all swapping constraints or because the language model score of the best candidate is below the threshold. The remaining pairs (16,280 for \qqp\ and 50k for Wikipedia) are passed to human review. 

\paragraph{Sentence correction} The examples generated using both of our strategies are generally of high quality, but they still need to be checked with respect to grammar and coherence. Annotators evaluate each generated sentence without seeing its source sentence. The sentence is accepted as is, fixed, or rejected. Table \ref{tb:swap_annotation} shows the number of pairs of each action on each domain. Most of fixes are minor grammar corrections like \textit{a apple$\rightarrow$an apple}. Accepted and fixed sentences are then passed to the next stage for paraphrase annotation. Overall 88\% of generated examples passed the human correction phase on both domains. 

\paragraph{Paraphrase identification} Sentence pairs are presented to five annotators, each of which gives a binary judgment as to whether they are paraphrases or not. We choose binary judgments to make our dataset have the same label schema as the \qqp\ corpus. Table \ref{tb:swap_annotation} shows aggregated annotation statistics on both domains, including the number of paraphrase (positive) and non-paraphrase (negative) pairs and human agreement, which is the percentage ratio of agreement between each individual label and the majority vote of five labels on each example pair. Overall, human agreement is high on both Quora (92.0\%) and Wikipedia (94.7\%) and each label only takes about 24 seconds. As such, answers are usually straightforward to human raters. 

To ensure the data is comprised of clearly paraphrase or non-paraphrase pairs, only examples with four or five raters agreeing are kept.\footnote{We exclude low agreement pairs from our experiments, but we include them in our data release for further study.} An example of low agreement is \textit{Why is the 20th-century music so different from the 21st music?} v.s. \textit{Why is the 21st century music so different from the 20th century music?}, where three out of five raters gave negative labels on this pair. The bottom block of Table \ref{tb:swap_annotation} shows the final number of pairs after this filtering, and human agreement further goes up to over 95\%. Finally, source and generated sentences are randomly flipped to mask their provenance. 

\begin{table}[t]
    \centering
    \begin{tabular}{lr}
         Total \# back translation pairs & 26,897 \\
         ~~~~paraphrase & 25,521 \\
         ~~~~non-paraphrase & 1,376 \\
         Human agreement & 94.8\% \\
    \end{tabular}
    \caption{Paraphrase judgments on example pairs generated by back translation on Wikipedia sentences.}
    \label{tb:bt_annotation}
\end{table}

The swapping strategy generally produces non-paraphrase examples---67\% for \qqp\ and 88\% for Wikipedia. Because (a) the label imbalance is less pronounced for \qqp\ and (b) NMT models perform poorly on Quora questions due to domain mismatch, we only apply the back translation strategy to Wikipedia pairs. Doing so creates 26,897 candidate example pairs after filtering. As before, each pair is rated by five annotators on the paraphrase identification task.\footnote{Sentence correction was not necessary for these because NMT generates fluent output.} Table \ref{tb:bt_annotation} shows that most of the examples (94.9\%) are paraphrases (as expected), with high human agreement (94.8\%). Finally, we expand the pairs using the the rules described in Section \ref{sec:backtrans}.

\begin{table}[t]
    \centering
    \begin{tabular}{l@{~~~}r@{~~~}r@{~~~}r@{~~~}r}
         & Train & Dev & Test & Yes\%\\
    \midrule
         \wsq & 11,988 & 677 & -- & 31.3\% \\
         \wsw & 49,401 & 8,000 & 8,000 & 44.2\% \\
         \wswswap & 30,397 & -- & -- & 9.6\% \\
    \end{tabular}
    \caption{Counts of experimental split for each \paws\ dataset. The final column gives the proportion of paraphrase (positive) pairs. There are 108,463 \paws\ pairs in total.}
    \label{tab:paws-stats}
\end{table}

Table \ref{tab:paws-stats} provides counts for each split in the final PAWS datasets. The training portion of \wsq\ is a subset of the \qqp\ training set; however, \wsq's development set is a subset of both \qqp's development and test sets because there are only 677 pairs. \wsw\ randomly draws 8,000 pairs for each of its development and test sets and takes the rest as its training set, with no overlap of source sentences across sets. Finally, any trivial pairs with identical sentences from development and test sets are removed.\footnote{Such trivial examples exist because annotators sometimes fix a swapped sentence back to its source. We keep such examples in the training set (about 8\% of the corpus) because otherwise a trained model would actually predict low similarity scores to identical pairs.}  The final \wsq\ has a total of 12,665 pairs (443k tokens), where 31.3\% of them have positive labels (paraphrases). \wsw\ has a total of 65,401 pairs (2.8m tokens), where 44.2\% of them are paraphrases.

Note that we have human annotations on 43k pairs generated by the word swapping method on Wikipedia, but 30k of them have no back translation counterparts and therefore they are not included in our final \wsw\ dataset. Nevertheless, they are high-quality pairs with manual labels, so we include them as an auxiliary training set (\wswswap\ in Table \ref{tab:paws-stats}), and empirically show its impact in Section \ref{sec:experiments}.

\paragraph{Unlabeled \wsw} In addition to the fully labeled \wsw\ dataset, we also construct an unlabeled \wsw\ set at large scale. The idea is to simply treat all pairs from word swapping as non-paraphrases and all pairs from back translation as paraphrase, and construct the dataset in the same way as labeled \wsw. The result is a total of 656k pairs with silver labels. We show empirically the impact of using this silver set in pre-training in Section \ref{sec:experiments}.

%% file: models.tex
\section{Evaluated Models}
\label{sec:models}

\paws\ is designed to probe models' ability to go beyond recognizing overall sentence similarity or relatedness. As noted in the introduction, models---even the best avaliable---trained on existing resources tend to classify any example with high BOW overlap as a paraphrase. Can any of these models learn finer structural sensitivity when provided with \paws\ examples as part of their training? 

\begin{table}[t]
    \centering
    \begin{tabular}{@{~}l@{~~}c@{~~~~}c@{~~~~}c@{~~~~}c@{~}}
      & \small{BOW} & \specialcell{\small{BiLSTM}\\\small{\& ESIM}} & \small{DecAtt} & \specialcell{\small{DIIN \&}\\\small{BERT}}  \\
    \midrule
    \small{Non-local context} & \No & \Yes & \No & \Yes \\
    \small{Word interaction} & \No & \No & \Yes & \Yes \\
    \end{tabular}
    \caption{Complexity of each evaluated model.}
    \label{tb:models}
\end{table}

We consider six different models that cover a wide range of complexity and expressiveness: two baseline encoders and four recent advanced models that achieved state-of-the-art or strong performance on paraphrase identification. Table \ref{tb:models} summarizes the models with respect to whether they represent non-local contexts or support cross-sentential word interaction. 

The baseline models use cosine similarity with simple sentence encoders: a bag-of-words (\textbf{BOW}) encoder based on token unigram and bigram encodings and a bi-directional LSTM (\textbf{BiLSTM}) that produces a contextualized sentence encoding. A cosine value above .5 is taken as a paraphrase.

\textbf{ESIM}. The Enhanced Sequential Inference Model \cite{ChenZLWJI17} achieved competitive performance on eight sentence pair modeling tasks \cite{lan018neural}. It encodes each sentence using a BiLSTM, concatenates the encodings for each sentence in the pair, and passes them through a multi-layer perceptron (MLP) for classification. The additional layers allow ESIM to capture more complex sentence interaction than cosine similarity in the baseline models.

\textbf{DecAtt}. The Decomposable Attention Model \cite{ParikhT0U16} is one of the earliest models to introduce attention for paraphrase identification. It computes word pair interaction between two sentences and aggregates aligned vectors for final classification. This model achieved state-of-the-art results without explicitly modeling word order. In our experiments, we show the limitations of this modeling choice on \paws\ pairs.

\textbf{DIIN}. The Densely Interactive Inference Network \cite{gong2018natural} adopts DenseNet \cite{HuangLMW17}, a 2-dimensional convolution architecture, to extract high-order word-by-word interaction between n-gram pairs. This model achieved state-of-the-art performance without relying on pre-trained deep contextualized representations like ELMo \cite{PetersNIGCLZ18}. It outperformed ESIM and DecAtt models by a large margin on both paraphrase identification and natural language inference tasks.

\textbf{BERT}. The Bidirectional Encoder Representations from Transformers \cite{devlin2018bert} recently obtained new state-of-the-art results on eleven natural language processing tasks, including pushing the GLUE benchmark to 80.4\% (7.6\% absolute improvement). BERT involves pre-training a Transformer encoder \cite{VaswaniSPUJGKP17} on a large corpus with over three billion words. This large network is then fine-tuned with just one additional output layer.

%% file: experiments.tex
\begin{table*}[t]
    \centering
    \begin{tabular}{lcccccc}
	\toprule
    \multirow{2}{*}{\textsc{Models}} & \multicolumn{2}{c}{~~~\small{\qqp$\rightarrow$\qqp}~~~} & \multicolumn{2}{c}{~~\small{\qqp$\rightarrow$\wsq}~~} & \multicolumn{2}{c}{\small{\qqp+\wsq$\rightarrow$\wsq}}\\
     & \small{~~~(Acc)} & \small{(AUC)} & \small{~~~(Acc)} & \small{(AUC)} & \small{~~~(Acc)} & \small{(AUC)} \\
    \midrule
    BOW & ~~~83.2 & 89.5 & ~~~29.0 & 27.1 & ~30.0 \small{(+1.0)} & 27.3 \small{(+0.2)}\\
    BiLSTM & ~~~86.3 & 91.6 & ~~~34.8 & \textbf{37.9} & ~~~57.6 \small{(+22.9)} & ~~52.3 \small{(+14.5)}\\
    ESIM \cite{ChenZLWJI17} & ~~~85.3 & 92.8 & ~~~\textbf{38.9} & 26.9 & ~~~66.5 \small{(+27.7)}& ~~48.1 \small{(+17.2)}\\
    DecAtt \cite{ParikhT0U16} & ~~~87.8 & 93.9 & ~~~33.3 & 26.3 & ~~~67.4 \small{(+34.1)} & ~~51.1 \small{(+24.9)}\\
    DIIN \cite{gong2018natural} & ~~~89.2 & 95.2 & ~~~32.8 & 32.4 & ~~~83.8 \small{(+51.1)} & ~~77.8 \small{(+45.5)}\\
    BERT \cite{devlin2018bert} & ~~~\textbf{90.5} & \textbf{96.3} & ~~~33.5 & 35.1 & ~~~\textbf{85.0} \small{(+51.5)} & ~~\textbf{83.1} \small{(+48.0)}\\
    \bottomrule
    \end{tabular}
    \caption{\textbf{Acc}uracy (\%) of classification and \textbf{AUC} scores (\%) of precision-recall curves on Quora Question Pairs (\textbf{\qqp}) testing set and our \wsq\ development set. \qqp$\rightarrow$\wsq\ indicates that models are trained on \qqp\ and evaluated on \wsq. Other columns are defined in a similar way. \qqp+\wsq\ is a simple concatenation of the two training sets. Boldface numbers indicate the best accuracy for each testing scenario. Numbers in parentheses indicate absolute gains from adding \wsq\ training data.}
	\label{tb:quora_results}
\end{table*}

\section{Experiments} \label{sec:experiments} 

We seek to understand how well models trained on standard datasets perform on \paws\ pairs and to see which models are most able to learn from \paws\ pairs. A strong model should improve significantly on \paws\ when trained on \paws\ pairs without diminishing performance on existing datasets like QQP. Overall, both DIIN and BERT prove remarkably able to adapt to \paws\ pairs and perform well on both \wsq\ and \wsw\ while the other models prove far less capable. 

\subsection{Experimental Setup}
We use two metrics: classification accuracy and area-under-curve (AUC) scores of precision-recall curves. For all classification models, 0.5 is the threshold used to compute accuracy. We report results on testing sets for QQP and \wsw, and on the development set for \wsq\ (which has no test set).

For BERT, we use the implementation provided by the authors\footnote{\scriptsize{\url{ https://github.com/google-research/bert}}} and apply their default fine-tuning configuration. We use the provided BERT$_{\textrm{BASE}}$ pre-trained model instead of BERT$_{\textrm{LARGE}}$ due to GPU memory limitations. For all other models, we use our own (re-)implementations that matched reported performance on \qqp. We use 300 dimensional GloVe embeddings \cite{PenningtonSM14} to represent words and fix them during training.

\subsection{Results}

\paragraph{Main Results on \wsq} Table \ref{tb:quora_results} summarizes results on the Quora domain. We first train models on the Quora Question Pairs (\qqp) training set, and column ``\qqp$\rightarrow$\qqp'' shows that all models achieve over 83\% accuracy on \qqp. However, when evaluating on \wsq, all models, including BERT, obtain abysmal accuracy under 40\% (column ``\qqp$\rightarrow$\wsq''). 

We hypothesize the performance on \wsq\ relies on two factors: the number of representative training examples, and the capability of models to represent complex interactions between words in each sentence and across the sentences in the pair. To verify that, we further train models on a combination of \qqp\ and \wsq\ training sets and the last two columns of Table \ref{tb:quora_results} show the results on \wsq. As expected, all models benefit from new training examples, but to different extents. Gains are much larger on state-of-the-art models like BERT, while the BOW model learns almost nothing from new examples. As a consequence, performance changes are more drastic on \wsq\ than on \qqp. For example, the absolute difference between BiLSTM and BERT is 4.2\% on \qqp, but it goes up to 27\% on \wsq, which is a 60\% relative reduction in error. 

It is also noteworthy that adding \wsq\ training examples has no negative impact to \qqp\ performance at all. For example, a BERT model fine-tuned on \qqp+\wsq\ achieves the same 90.5\% classification accuracy as training on \qqp\ alone. We therefore obtain a single model that performs well on both datasets.

\begin{table}[t]
    \centering
    \begin{tabular}{lcccc}
	\toprule
    \multirow{2}{*}{\textsc{Models}} & \multicolumn{2}{c}{\small{~~~Supervised~~~}} & \multicolumn{2}{c}{\small{Pretrain+Fine-tune}}\\
    & \small{(Acc)} & \small{(AUC)} & \small{~~(Acc)} & \small{(AUC)} \\
    \midrule
    BOW & 55.8 & 41.1 & 55.6 & 44.9\\
    BiLSTM & 71.1 & 75.6 & 80.8 & 87.6\\
    ESIM & 67.2 & 69.6 & 81.9 & 85.8\\
    DecAtt & 57.1 & 52.6 & 55.8 & 45.4\\
    ~~~+BiLSTM & 68.6 & 70.6 & 88.8 & 92.3\\
    DIIN & 88.6 & 91.1 & 91.8 & \textbf{94.4}\\
    BERT & \textbf{90.4} & \textbf{93.7} & \textbf{91.9} & 94.3\\
    \bottomrule
    \end{tabular}
    \caption{Accuracy (\%) and AUC scores (\%) of different models on \wsw\ testing set. \textbf{Supervised} models are trained on human-labeled data only, while \textbf{Pretrain+Fine-tune} models are first trained on noisy unlabeled \wsw\ data and then fine-tuned on human-labeled data.}
	\label{tb:wiki_results}
\end{table}

\paragraph{Main Results on \wsw} In our second experiment we train and evaluate models on our \wsw\ dataset. Table \ref{tb:wiki_results} presents the results. DIIN and BERT outperform others by a substantial margin ($>$17\% accuracy gains). This observation gives more evidence that \paws\ data effectively measures models' sensitivity to word order and syntactic structure.

One interesting observation is that DecAtt performs as poorly as BOW on this dataset. This is likely due to the fact that DecAtt and BOW both consider only local context information. We therefore tested an enhancement of DecAtt by replacing its word representations with encodings from a BiLSTM encoder to capture non-local context information. The enhanced model significantly outperforms the base, yielding an 11.5\% (57.1\% vs. 68.6\%) absolute gain on accuracy.

We further evaluate the impact of using silver \wsw\ data in pre-training, as discussed in Section \ref{sec:ws_dataset}. The last two columns of Table \ref{tb:wiki_results} show the results. Comparing to supervised performance, pre-training with silver data gives consistent improvements across all models except BOW and vanilla DecAtt. Perhaps surprisingly, adding silver data gives more than 10\% absolute improvements on AUC scores for BiLSTM and ESIM, much higher than the gains on DIIN and BERT.

\begin{figure}[t]
\centering
\includegraphics[width=0.47\textwidth]{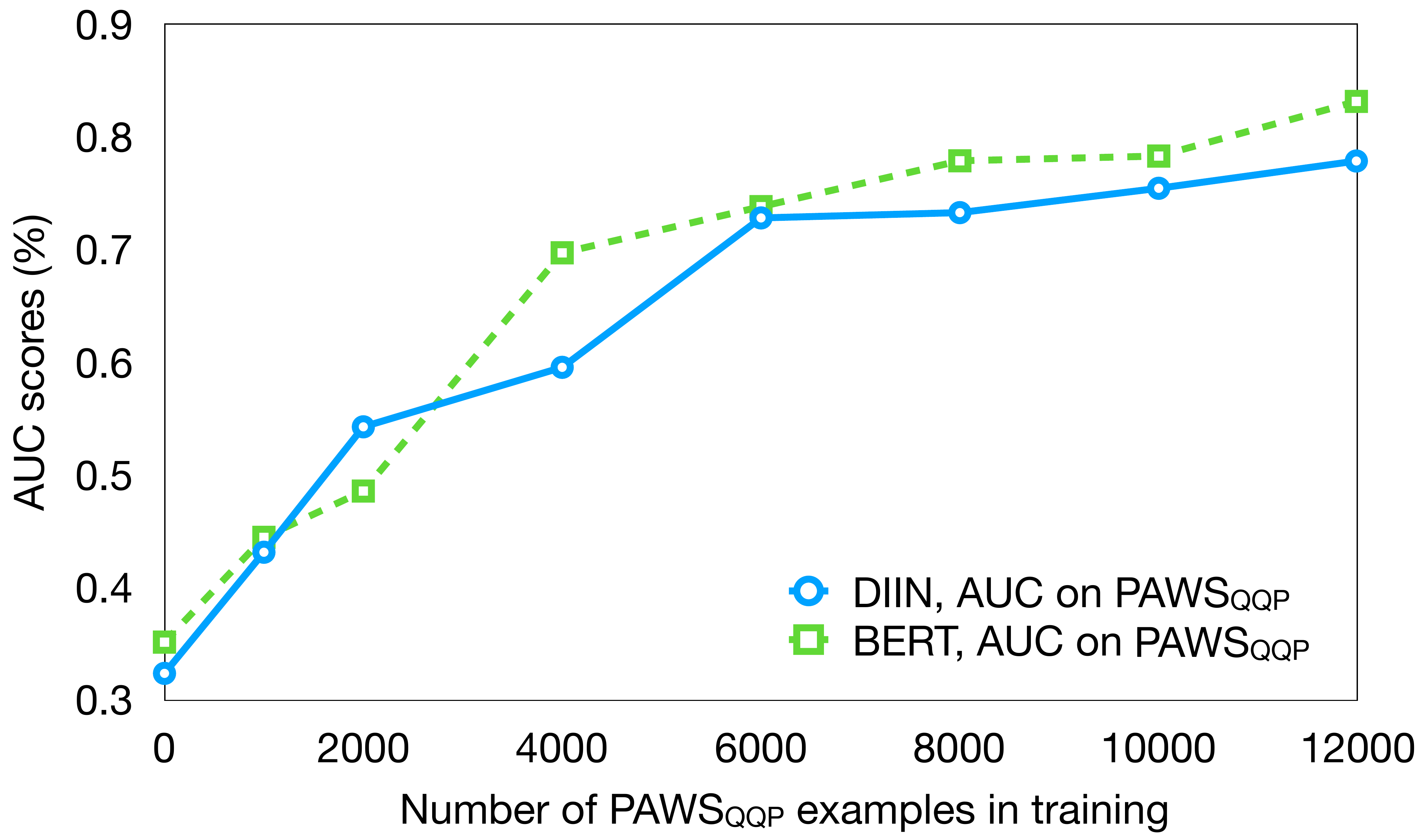}
\caption{AUC scores ($y$-axis) as a function of the number of \wsq\ examples in the training set ($x$-axis).}\label{fig:learning_curve}
\end{figure}

\paragraph{Size of Training Set} To analyze how many \paws\ examples are sufficient for training, we train multiple models on \qqp\ plus different number of \wsq\ examples. Figure \ref{fig:learning_curve} plots AUC score curves of DIIN and BERT as a function of the number of \wsq\ training examples. $x=0$ corresponds to models trained on \qqp\ only, and the rightmost points correspond to models trained on \qqp\ and full \wsq. Both models improve from 30\% to 74\% AUC scores with 6,000 \wsq\ examples. Furthermore, neither curve reaches convergence, so they would likely still benefit from more \paws\ training examples.

\paragraph{Cross-domain Results} The \paws\ datasets cover two domains: Quora and Wikipedia. Here we demonstrate that a model trained on one domain also generalizes to another domain, although not as well as training on in-domain data. Table \ref{tb:training_sets} shows that a DIIN model trained on Quora (\qqp+\wsq) achieves 70.5\% AUC on the Wikipedia domain. This is lower than training on in-domain data (92.9\%), but higher than the model trained without any \paws\ data (46.0\%). We also observe similar patterns when training on Wikipedia (\qqp+\wsw) and testing on \wsq. Interestingly, using out-of-domain data also boosts in-domain performance. As Table \ref{tb:training_sets} shows, training on both domains (\qqp+\wsqw) leads to 9.2\% absolute AUC gains on \wsq\ over the model trained only on \qqp+\wsq.

The auxiliary training set on Wikipedia (\wswswap) helps further. As Table \ref{tb:training_sets} shows, adding this auxiliary training set is particularly helpful to the performance on \wsq, yielding a 12.1\% (70.6\% vs 58.5\%) gain on AUC when training on \qqp+\wsw. On \wsw, this addition lifts the (no pre-training) DIIN model AUC from 91.1\% (Table \ref{tb:wiki_results}) to 93.8\% (Table \ref{tb:training_sets}).

\begin{table}[t]
    \centering
    \begin{tabular}{lcc@{~~}c}
	\toprule
	\multirow{2}{*}{\textsc{Training Data}} & \small{\qqp}  & \small{\wsq} & \small{\wsw} \\
	& \small{(Test)} & \small{(Dev)} & \small{(Test)} \\
    \midrule
    \small{\qqp\ (Train)} & 95.2 & 32.4 & 46.0\\
    \small{\qqp+\wsq} & \textbf{95.3} & 77.8 & 70.5\\
    \small{\qqp+\wsw} & \textbf{95.3} & 58.5 & 92.9\\
    \small{~~~~~+\wsw$_{\textrm{-Swap}}$} & \textbf{95.3} & 70.6 & 93.5\\
    \small{\qqp+\wsqw} & 95.1 & 87.0 & 93.4\\
    \small{~~~~~+\wsw$_{\textrm{-Swap}}$} & \textbf{95.3} & \textbf{89.9} & \textbf{93.8}\\
    \bottomrule
    \end{tabular}
    \caption{AUC scores (\%) when training DIIN models on different sets of training data. Boldface numbers indicate the best accuracy for each testing set. }
	\label{tb:training_sets}
\end{table}

\paragraph{BERT vs DIIN} Both models achieve top scores on \paws, but interestingly, the two models disagree on many pairs and are not correlated in their errors. For example, of 687 of BERT's mistakes on the \wsw\ test set,  DIIN got 280 (41\%) correct. As such, performance might improve with combinations of these two existing models.

It is also worth noting that the DIIN model used in our experiments has only 590k model parameters, whereas BERT has over 100m. Furthermore, the computational cost of BERT is notably higher than DIIN. Given this, and the fact that DIIN is competitive with BERT (especially when pre-trained on noisy pairs, see Table \ref{tb:wiki_results}), DIIN is likely the better choice in computationally constrained scenarios---especially those with strict latency requirements.

%% file: conclusion.tex
\section{Conclusion}
\label{sec:conclusion}

Datasets are insufficient for differentiating models if they lack examples that exhibit the necessary diagnostic phenomena. This has led, for example, to new datasets for noun-verb ambiguity \cite{NOUNVERB} and gender bias in coreference \cite{webster2018gap,rudinger-EtAl:2018:N18-2,ZWYOC18}. 
Our new \paws\ datasets join these efforts and provide a new resource for training and evaluating paraphrase identifiers. We show that including PAWS training data for state-of-the-art models dramatically improves their performance on challenging examples and makes them more robust to real world examples. We also demonstrate that PAWS effectively measures sensitivity of models on word order and syntactic structure.